\documentclass[conference]{IEEEtran}
\IEEEoverridecommandlockouts
\usepackage{cite}
\usepackage{amsmath,amssymb,amsfonts}
\usepackage{graphicx}
\usepackage{booktabs}
\usepackage{textcomp}
\usepackage{xcolor}
\usepackage[nothing]{algorithm}
\usepackage{algorithmic}
\usepackage{caption}
\usepackage{subcaption}
\usepackage{hyperref}

\def\BibTeX{{\rm B\kern-.05em{\sc i\kern-.025em b}\kern-.08em
    T\kern-.1667em\lower.7ex\hbox{E}\kern-.125emX}}
\begin{document}

\title{Exploring open-ended gameplay features with Micro RollerCoaster Tycoon\\
\thanks{This work was supported by the National Science Foundation (Award number 1717324 - RI: Small: General Intelligence through Algorithm Invention and Selection).}
}

\author{\IEEEauthorblockN{Michael Cerny Green}
\IEEEauthorblockA{\textit{OriGen.AI, Tandon School of Engineering} \\
\textit{New York University}\\
Brooklyn, USA \\
mike.green@nyu.edu
}
\and
\IEEEauthorblockN{Victoria Yen}
\IEEEauthorblockA{\textit{Tandon School of Engineering} \\
\textit{New York University}\\
Brooklyn, USA \\
victoria.yen@nyu.edu}
\and
\IEEEauthorblockN{Sam Earle}
\IEEEauthorblockA{\textit{Tandon School of Engineering} \\
\textit{New York University}\\
Brooklyn, USA\\
sam.earle@nyu.edu}
\and
\IEEEauthorblockN{Dipika Rajesh}
\IEEEauthorblockA{\textit{Independent Researcher} \\ \\
Brooklyn, USA \\ 
dipika.rajesh@gmail.com}
\and
\IEEEauthorblockN{Maria Edwards}
\IEEEauthorblockA{\textit{Tandon School of Engineering} \\
\textit{New York University}\\
Brooklyn, USA \\
mariaedwards@nyu.edu}
\and
\IEEEauthorblockN{L. B. Soros}
\IEEEauthorblockA{\textit{Cross Labs} \\
\textit{Cross Compass, Ltd.}\\
Tokyo, Japan \\
lisa.soros@cross-compass.com}
}

\maketitle

\begin{abstract}
This paper introduces MicroRCT, a novel open source simulator inspired by the theme park sandbox game RollerCoaster Tycoon. The goal in MicroRCT is to place rides and shops in an amusement park to maximize profit earned from park guests. Thus, the challenges for game AI include both selecting high-earning attractions and placing them in locations that are convenient to guests. In this paper, the MAP-Elites algorithm is used to generate a diversity of park layouts, exploring two theoretical questions about evolutionary algorithms and game design: 1) Is there a benefit to starting from a minimal starting point for evolution and complexifying incrementally? and 2) What are the effects of resource limitations on creativity and optimization? Results indicate that building from scratch with no costs results in the widest diversity of high-performing designs.
\end{abstract}


\section{Introduction}

RollerCoaster Tycoon is a sandbox / management simulation game first developed by Chris Sawyer in 1999\footnote{Amazingly, Sawyer programmed 99\% of the original game in x86 assembler/machine code. \cite{sawyerFAQ}}. The goal is to build a successful amusement park containing attractions such as rides and shops in addition to paths connecting the attractions. As the park grows, it will attract guest agents called ``peeps'' that are not controlled by the player, but are instead determined by simple heuristics and pathfinding algorithms. A successful park will contain hundreds or potentially thousands of guests that each spend money in the park. Winning the game thus entails achieving some target profit level after a certain amount of time has passed.

Sandbox games such as RollerCoaster Tycoon are interesting because they afford open-ended gameplay; there are many ways to play, and in this case, many ways to win. This paradigm is in contrast to closed games such as platformers, for instance, where only a very restricted set of possible paths through a level will allow the player to win the game. In the context of automated tools for game design, an important challenge is designing algorithms that can illuminate the space of strategies afforded by a particular design configuration. That way, the game parameters and mechanics can be adjusted so that a maximal number of viable strategies are afforded. 

\begin{figure}[t]
\center{
    \includegraphics[width=0.6\linewidth]{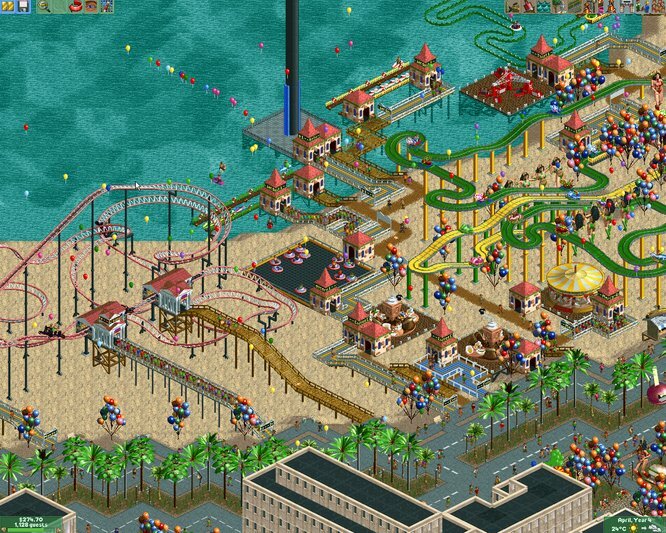}
    }
\caption{\textbf{A screenshot from RollerCoaster Tycoon 2.} The open source MicroRCT environment introduced in this paper abstracts key features of this game into a minimal environment for testing generative design algorithms.}
\label{fig:rct}
\end{figure}

This paper uses the MAP-Elites algorithm \cite{cully:nature15} to illuminate the design space afforded by two sandbox game features. The first question addressed in this paper is whether there is any benefit to starting from a completely blank park ``canvas'' as opposed to seeding the design space with some starter attractions. The second question addressed in this paper is how the diversity of viable strategies is affected by resource limitations. In RollerCoaster Tycoon in particular, possible parks are limited by the amount of simulated money the player has accumulated at any given time because each attraction and path tile has a cost. In some levels, this resource limitation is mitigated by giving the player infinite money, enabling a subtly different kind of gameplay experience.

\section{Background}

\subsection{The MAP-Elites Algorithm}

MAP-Elites \cite{cully:nature15} is a canonical quality-diversity algorithm \cite{pugh:frontiers16}. This class of algorithms, which is commonly used for optimization and procedural generation in robotics and games \cite{QDio, khalifa:gecco18, gravina2019procedural, alvarez2019empowering,charity2020mech}, combines a standard evolutionary search algorithm with a diversity-preserving archive or map that is used as a container for diverse elites in the population \cite{cully:ToEC17}. The archive in MAP-Elites in particular takes the form of a grid of discrete cells. By enforcing diversity preservation while the search algorithm is running, the algorithm can uncover a variety of high-performing individuals in some search space. Besides MAP-Elites, the other canonical quality-diversity algorithm is novelty search with local competition \cite{lehman:gecco2011}. 

Most quality-diversity algorithms require the programmer to specify a feature vector, also called a behavior characterization or behavioral descriptor, to parameterize the archive of elites. That is, the user must manually specify along which dimensions they'd like to see diversity. Notable exceptions to this claim include AURORA \cite{cully:gecco19}, which automatically defines a two-dimensional feature vector using principal component analysis, and meta-evolution \cite{bossens:gecco20}, which leverages covariance matrix adaptation for the same purpose. 


\subsection{AI for simulation games}
A few works have focused on experiments with the simulation game Transport Tycoon \cite{rios:sb09,wisniewski:thesis,bijlsma:thesis,lakomy:thesis,katreniakova:thesis,konijnendijk:thesis}, which was a predecessor of RollerCoaster Tycoon (and was also developed by Chris Sawyer). In this game, the primary task is to maximize profit earned by a resource transportation system. Other works have investigated AI for playing SimCity \cite{earle2020using}. The goal in this game is to manage the placement of commercial, industrial, and residential zones (along with utilities such as roads and power lines) to maximize population growth in a city simulated by a cellular automata.  

Notably, the studies on Transport Tycoon and SimCity mentioned in the prior paragraph all focused on learning to play these design-based games \emph{well}, but did not focus explicitly on exploration of the design space available to players (automated or otherwise) playing the game. Charity et al.\ address this topic directly in their work on SimSim, a testbed for sandbox AI that is inspired by interior design in The Sims \cite{charity:aiide20}. In this work, the authors explore the ability of a minimal criterion evolutionary algorithm \cite{soros:alife14,brant:gecco17} to illuminate the design space. In the minimal criterion paradigm, which is expressly distinct from optimization, all individuals that meet some performance threshold are considered viable (and are thereby allowed to reproduce), while all others are considered nonviable. The goal in this study was to maximize the number of individuals that satisfied a minimal viable design criterion, but not to achieve any specific objectives otherwise. In contrast, the current study in RollerCoaster Tycoon aims to maximize design diversity in a domain that \emph{is} winnable by achieving a target profit level. 

\section{MicroRCT}

\begin{figure}[t]
\center{
    \includegraphics[width=0.5\linewidth]{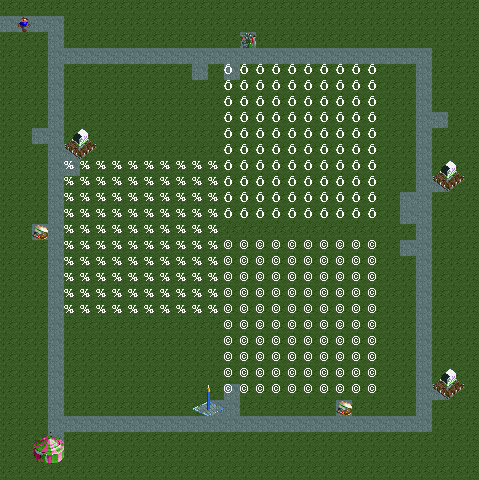}
    }
\caption{\textbf{A screenshot from the MicroRCT environment.} In this simplified environment, the task is to place attractions along a fixed, donut-shaped path. Roller coasters are represented by ASCII characters, while smaller atrractions are rendered with their own graphics (from the freely available openRCT2 project). The park layout is decided before guests enter the park and then cannot be modified during evalutation. Path stubs are leftover from attractions that have been placed and then removed during the evolutionary process.}
\label{fig:micro}
\end{figure}

For the purpose of experiments in this paper, a simplified version of RollerCoaster Tycoon (RCT) called MicroRCT (Figure \ref{fig:micro})\footnote{Publicly available at \url{https://github.com/smearle/micro-rct}} was created. It is based on code from the OpenRCT2 framework\footnote{Available at \url{https://github.com/OpenRCT2/OpenRCT2}}. This section addresses which features from the original game are included in MicroRCT and highlights significant differences.

To allow for a wide diversity of AI-generated designs, a variety of rides and paths is included within a park. In the original RCT, the park can be expanded and paths can be added or removed freely, but this feature was removed for ease of analysis. Instead, in MicroRCT the park size and the main pathway are fixed and gameplay is focused only on the placement of rides, which includes roller coasters, restrooms, first aid buildings, cinemas, transportation, and shops. 

\begin{table}[]
\begin{tabular}{@{}llllll@{}}
\toprule
Ride Name                & Excitement & Intensity & Nausea & Cost/tile & Char \\ \midrule
Bobsleigh Coaster        & 50         & 30        & 10     & 45            & \&     \\
Corkscrew R.C. & 50         & 30        & 10     & 50            & \%     \\
Junior R.C.    & 50         & 30        & 10     & 40            & j      \\
Reverser R.C.  & 48         & 28        & 7      & 37            & $\ll$      \\
Wooden R.C.    & 52         & 33        & 8      & 50            & W      \\
Car Ride                 & 70         & 10        & 10     & 25            & æ      \\
Circus                   & 20         & 10        & 0      & 125           & c      \\
Crooked House            & 15         & 8         & 0      & 65            & /      \\
Observation Tower        & 80         & 10        & 0      & 57            & T      \\
Spiral Slide             & 50         & 10        & 0      & 165           & §      \\
Cinema 3D                & 20         & 10        & 0      & 140           & E      \\
Launched Freefall        & 50         & 50        & 10     & 70            & F      \\
Magic Carpet             & 50         & 30        & 10     & 198           & $\sim$ \\
Motion Simulator         & 20         & 10        & 10     & 220           & ¶      \\
Twist                    & 40         & 20        & 10     & 90            & ¥      \\
Boat Hire                & 70         & 6         & 0      & 55            & B      \\
Dinghy Slide             & 50         & 30        & 10     & 40            & d      \\
Log Flume                & 80         & 30        & 6      & 40            & Õ      \\
Splash Boats             & 80         & 34        & 6      & 30            & v      \\
Submarine Ride           & 70         & 6         & 0      & 70            & ©      \\
Chairlift                & 70         & 10        & 0      & 30            & ¯      \\
Lift                     & 80         & 10        & 0      & 39            & L      \\
Miniature Railway        & 70         & 6         & -10    & 35            & r      \\
Suspended Monorail       & 70         & 6         & -10    & 50            & M      \\ \bottomrule
\end{tabular}

\caption{\textbf{List of rides in the MicroRCT environment.} ``R.C.'' stands for ``roller coaster'' in ride names.}
\label{table:vals}
\end{table}

Park guests, called ``peeps'' have several basic parameters. At the initial state each peep has a happiness value between 113 to 144 (max value is 256). Each guest has another important parameter called the ``happiness target'', in the original RCT code. After rides and purchases change the happiness level, happiness is updated toward the happiness target on each simulation tick. The main interactions between rides and guests in RCT include making purchases, using the facilities (toilet and first aid), riding rides, and queuing for attractions. Queuing was not included in MicroRCT because it requires agents to consider the queue’s size in the physical space, which makes ride selection calculations unnecessarily more difficult.

In the original RCT, parks start with zero peeps and attract new peeps as the result of advertising promotions and the increasing reputation of the park. In MicroRCT, this game mechanic is simplified by setting a fixed amount of guests inside the park. Guests in the experiments stay in the park for the duration of the evaluation and continuously interact with rides and shops. 

Each type of ride has different settings (excitement, intensity, and nausea) for different instances of the type of ride. In the original RCT, players can create custom roller coaster designs, resulting in various physical sizes, excitement values, intensity values, and nausea values for these rides. These parameters can affect how happy the rides make peeps. In MicroRCT, every roller coaster has the same basic parameterization and thus has the same effect on peeps. 

Another simplification is that peeps' interactions with rides are nearly instantaneous, so it only takes one tick for guests to interact with rides and update their personal parameters. There are several parameters that are affected, and happiness, nausea and happiness target are the most relevant. Happiness target is calculated by considering each peep’s tolerance toward the intensity and nausea. After the happiness target changes, happiness will move toward the happiness target.

In the original RCT, peeps will either consider all the rides or only consider the nearest ride. After these rides are put into consideration peeps will filter through several checks, such as leaving the park or not, checking to see if it is raining, etc. In MicroRCT, peeps only consider whether the nearest ride is too intense for their preferences. After deciding on the goal attraction, our peep will use DFS to navigate to their goal, which is the same as in original RCT. As in RCT, MicroRCT also includes a vomit mechanism whereby if a peep's nausea level exceeds 128 (255 is the maximum), they have a chance to throw up on the ground, and if the environment gets too dirty peeps’ happiness targets drop. Peeps will also try to find the first aid when nausea level is more than 128. 

\section{Methods}
The MAP-Elites evolutionary algorithm is used to explore MicroRCT design space. Algorithm \ref{fig:psuedocode} gives psuedocode: 

\begin{algorithm}
\caption{MAP-Elites Algorithm}\label{fig:psuedocode}
\begin{algorithmic}[1]
\STATE Initialize population of size $n$ parks
\FOR{$g$ number of generations}
\STATE Simulate entire population for $t$ number of gameticks
\STATE Calculate fitness and dimensions of all parks
\FOR{each park in the population}
\IF{Elite map cell is empty}
\STATE Populate the cell with this park
\ELSIF{Elite map cell is full and park.fitness $>$ elite.fitness}
\STATE Replace elite map cell with park
\ELSE
\STATE Decide with probability $p$ to store this park within the elite cell's secondary population
\ENDIF
\ENDFOR
\STATE Randomly reinitialize generation from elite map
\ENDFOR
\end{algorithmic}
\end{algorithm}

A chromosome consists of a single park containing attractions, as well as a fitness score and dimensional metrics (see Section \ref{sec:methods-metrics}). Every park also contains a fixed circular path. Figure \ref{fig:micro} displays this main path, which always connects to the park entrance in the upper left corner of the park. Attractions are placed via mutation, and then depth-first search \cite{russell2002artificial} is used to find and build the shortest path between the attraction and the main park path. To mutate, a park undergoes $m$ number of changes to its layout. A change could mean adding a new attraction, deleting an existing attraction, or both (removing an existing attraction and replacing it with a new one). Adding new attractions may occur anywhere on the map, so long as the attraction will not overwrite forbidden areas, such as the main path.

Recall that MAP-Elites maintains a discretized grid of elites. Each grid cell may be populated with an elite chromosome, i.e. the chromosome that has the best fitness and matches the cell's dimensions. In the implementation in this paper in particular, the cell may also contain a secondary population, which may used to generate new chromosomes and maintain genetic diversity. Previous work showed that such \emph{passive genetic diversity} can improve performance \cite{pugh:thesis}.


\subsection{Dimension and Fitness Metrics}\label{sec:methods-metrics}

MicroRCT contains several metrics suitable for use as dimensions for MAP-Elites behavior vectors out-of-the-box. They can be organized into \emph{static} and \emph{runtime} metrics:

\subsubsection{Static}
Static metrics can be calculated at any time, as they are not reliant on park simulation:
\begin{itemize}
    \item Nausea: average nausea score of the park's attractions
    \item Excitement: average excitement score of attractions
    \item Intensity: average intensity score of attractions
    \item Diversity: Shannon entropy of attractions
\end{itemize}

\subsubsection{Runtime}
Runtime metrics can only be captured after park simulation as they are reliant on peep behavior:
\begin{itemize}
    \item Happiness: the average happiness of all peeps
    \item Vomit: the total number of times peeps have vomited
    \item Revenue: the total number of dollars peeps have spent
\end{itemize}

Attraction intensity, nausea and excitement scores are calculated by taking the mean of these features over all attractions in the park. (These per-attraction features are derived from the scores of corresponding attractions in the original game. The scores of various roller coaster types are based on the average of these scores among the set of pre-built track configuration that come with the game.) Similarly, guest happiness corresponds to the average happiness over all guests, calculated at the final timestep of simulation.

To calculate attraction diversity, each distinct attraction type is treated as a unique ``species,'' and Shannon entropy is used to compute the diversity over the ``population'' of all attraction instances present in the park. Shannon entropy is a popular choice for calculating the diversity of discrete populations, in both biology and evolutionary computation, and captures both the frequency and spread of and among a set of species.


Park income is calculated by taking the sum of money spent by guests on attractions and shop items in the park over the course of simulation. Vomit is calculated by counting the number of times guests have vomited in the park over the course of the simulation.

\section{Experiments}\label{sec:experiments}

The purpose of experiments in this paper, besides demonstrating the novel MicroRCT domain, is to explore how MAP-Elites can illuminate design space in a sandbox simulation game. In particular, this work focuses on two features that may enable open-ended gameplay: 1) constructing from scratch versus starting from a partial construction (\emph{initial park size and growth}), and 2) imposing cost constraints versus not (\emph{attraction cost constraint}). There is a binary option for park size/growth, a binary option for cost constraint, and four dimension combinations. Altogether, there are twelve experimental setups ($2$ x $2$ x $3$). Each setup is run $20$ times with different random seeds, for a total of $240$ elite maps. Each experimental run lasts $10,000$ generations. All parks are given $30$ x $30$ tiles in which attractions can be placed.


\subsection{Initial Park Size and Growth}
When generating the initial population, the MAP-Elites implementation uses a random park generator, which allows for the specification of the maximum number of attractions initially placed in a park in the first generation's initialization. There are two initial park sizes. A \emph{small} park is initialized with zero to four attractions. A \emph{medium} park is initialized with eight to twelve attractions. These ranges were chosen after preliminary experiments varying attraction counts and the park space available to place attractions. A park's initial size determines its mutation method. A small park has a 1 in 27 chance to remove a ride during mutation, and a $26$ (one for each attraction type) in $27$ chance to add an attraction (or replace an existing one in the park if this is selected and there is no space). A medium park has an equal chance to add an attraction or remove one.

\subsection{Attraction Cost Constraint}
Several experiments explored a monetary budget constraint for the MAP-Elites implementation. Attraction costs are listed in Table \ref{table:vals}. If the budget constraint is enabled, the algorithm is restricted from placing a set of attractions that exceeds this budget. Otherwise, the algorithm may select any set of attractions for the park, regardless of cost. In this work, the budget was set to $\$15,000$ if enabled, which was chosen after preliminary experimentation.

\subsection{MAP-Elites Behavior Dimensions}
Several combinations were selected from the metrics mentioned in Section \ref{sec:methods-metrics}, based on the authors' interest in how they might illuminate the game space:
\begin{itemize}
    \item Excitement \& Intensity
    \item Happiness \& Ride Diversity
    \item Happiness \& Vomit
\end{itemize}

Behavioral features are rounded down into bucket sizes, chosen based on preliminary experiments. For example, if Dimension A has bucket size 5, then any score from 0-4 would be rounded to 0, score 5-9 would rounded to 5, and so on. Excitement, Intensity, and Ride Diversity all have bucket size $5$. Happiness and Vomit both have bucket size $10$.

\section{Results}
Though this section contains only highlights of visual results, full experimental results have been made publicly accessible.~\footnote{ \url{drive.google.com/drive/folders/1ccIqqBjoVWlrN1chf1KaMQ9XbLhdJeY_}} To better consolidate results, all $20$ runs of each experiment permutation have been aggregated into a single hyperelite map, which contains the best elite of every map cell. All individual runs, elite maps, their programmatic representations, and aggregated elite maps can be viewed in the drive folder. The aggregated elite maps are presented in the following sections, along with their Quality-Diversity (QD) score. QD is calculated by taking the average fitness across the entire map.

\subsection{Costs}
\begin{figure}
     \centering
     \begin{subfigure}[b]{0.45\linewidth}
         \centering
         \includegraphics[width=\textwidth]{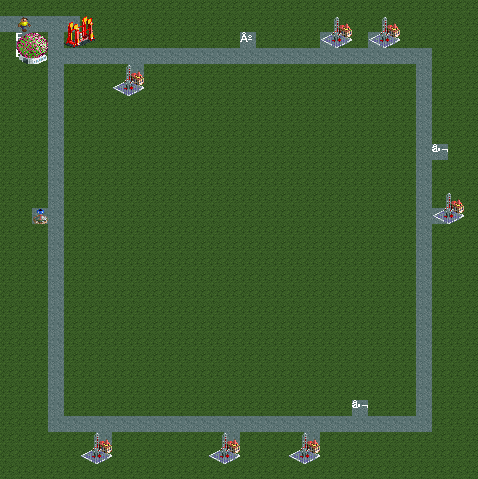}
         \caption{Cost enabled}
         \label{fig:cost-example}
     \end{subfigure}
     \hfill
     \begin{subfigure}[b]{0.45\linewidth}
         \centering
         \includegraphics[width=\textwidth]{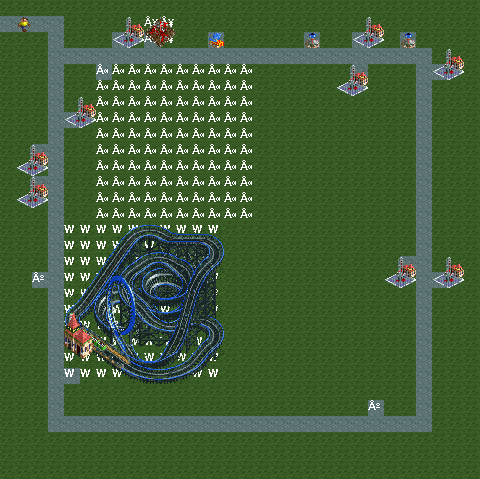}
         \caption{Cost disabled}
         \label{fig:nocost-example}
     \end{subfigure}
    \caption{\textbf{Two parks from excitement-intensity experiments}, both with excitement = 30 and intensity = 30. (a) contains freefall thrill rides with a fitness of $\$32,058$. Meanwhile, (b) contains two roller-coasters and freefall thrill rides, with a fitness of $\$38,160$}
    \label{fig:cost-nocost-example}
\end{figure}

To measure the effect of costs on evolution, only elite maps with the same dimensions and sizes are compared to one another. For example, small excitement-intensity with cost disabled maps are compared to small excitement-intensity elite maps with cost enabled. Generally, parks that do not have cost enabled have significantly higher fitness ceilings, where fitness is profit, than parks that take attraction costs into account. For example, for medium parks with excitement-intensity dimensions, the park with the highest fitness over $20$ runs is $\$40,000$, compared to the $\$55,000$ when cost is not considered. This pattern is seen across all tested dimension combinations and park sizes. This result is likely due to the fact that roller coasters, which generate more revenue, are more expensive to build than cheaper thrill rides. Figure \ref{fig:cost-nocost-example} supports this theory, displaying two parks with the same dimensions, different fitness values, and varying on whether or not cost is enabled. The cost-disabled park contains two large roller-coasters and freefall rides, while the cost-enabled park contains only freefall rides. Evidently, the cost of the roller-coasters is too high for the cost-enabled park. The cost-disabled park makes over $\$5,000$ more than its counterpart.

\subsection{Initial Size and Growth}\label{sec:results::size_and_growth}
To measure the effect of initialization sizes and growth methods on evolution, only elite maps with the same dimensions and cost-settings are compared to one another. For example, small excitement-intensity with cost disabled are compared to medium excitement-intensity elite maps with cost disabled. Initial size and growth methods do not seem to strongly affect fitness ceilings or floors of the elite maps. However, medium initialized park experiments seem to have a lower QD score than small counterparts for almost all experiments (the exception being happiness-vomit cost enabled parks). Although they may use different initialization sizes and growth methods, generators can still find similar park spaces. Figure \ref{fig:excitement_intensity_bothhigh_small} displays two elites, with opposite initialization and growth methods. Interestingly, they both found extremely similar parks with the same dimensional values.

\subsection{Excitement Intensity}
\begin{figure}
     \centering
     \begin{subfigure}[b]{0.45\linewidth}
         \centering
         \includegraphics[width=\linewidth]{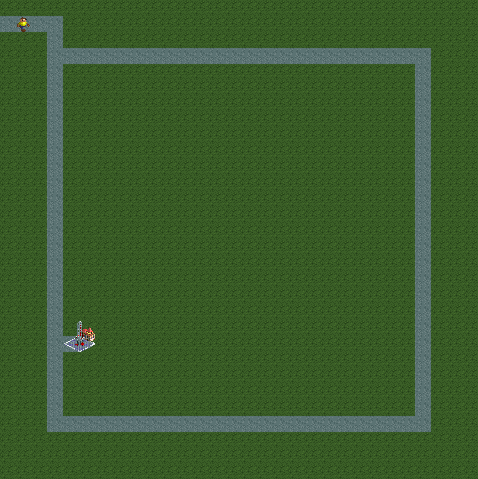}
        \caption{Small initialized park}
        \label{fig:excitement_intensity_bothhigh_small}
     \end{subfigure}
     \hfill
     \begin{subfigure}[b]{0.45\linewidth}
         \centering
         \includegraphics[width=\textwidth]{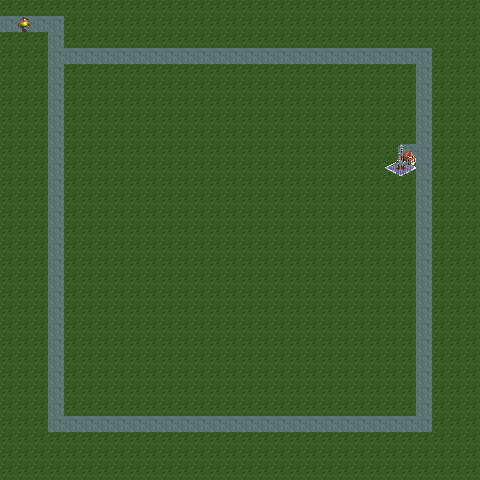}
         \caption{Medium initialized park}
         \label{fig:excitement_intensity_bothhigh_med}
     \end{subfigure}
    \caption{\textbf{Two parks from excitement-intensity experiments with different initialization and growth methods}, both with excitement = 50 and intensity = 50, containing the same ride. Though initialized and grown differently, the parks ended up being essentially identical. Both parks have a fitness value of roughly $\$15,000$.}
    \label{fig:excitement_intensity_bothhigh}
\end{figure}

Excitement and Intensity are both static metrics, meaning that they can be calculated independently of simulating the park, as they are based purely on the attraction metadata. As a result, parks using these dimensions have elite maps which are constrained by what is mathematically possible to produce. This can be seen in all four aggregated elite maps which have numerous empty cells for a variety of intensity and excitement score ranges. Parks with low excitement and intensity tend to have higher fitness scores than parks with incredibly high excitement and intensity. Figure \ref{fig:excitement_intensity_bothhigh} displays small and medium initialized parks, which both only contain one ride (which happens to have intensity and excitement scores of 50).

\begin{figure*}
     \centering
     \begin{subfigure}[b]{0.45\linewidth}
         \centering
         \includegraphics[width=\linewidth]{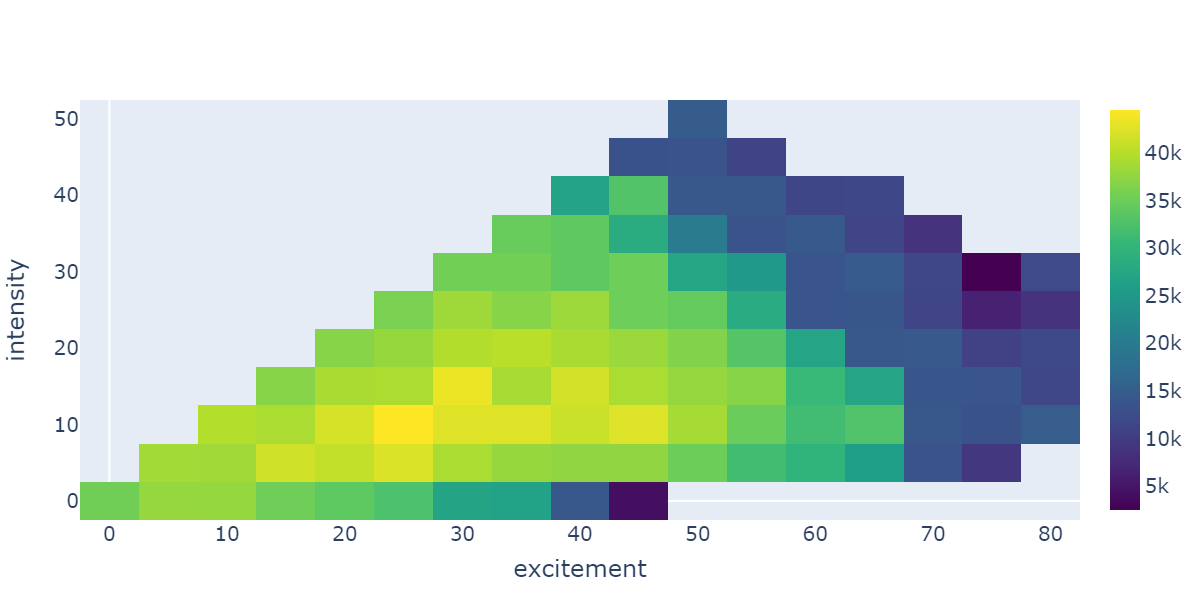}
        \caption{Small initialized, cost enabled park. QD: $\$27,801.56$}
        \label{fig:excitement_intensity_small_cost}
     \end{subfigure}
     \hfill
     \begin{subfigure}[b]{0.45\linewidth}
         \centering
         \includegraphics[width=\textwidth]{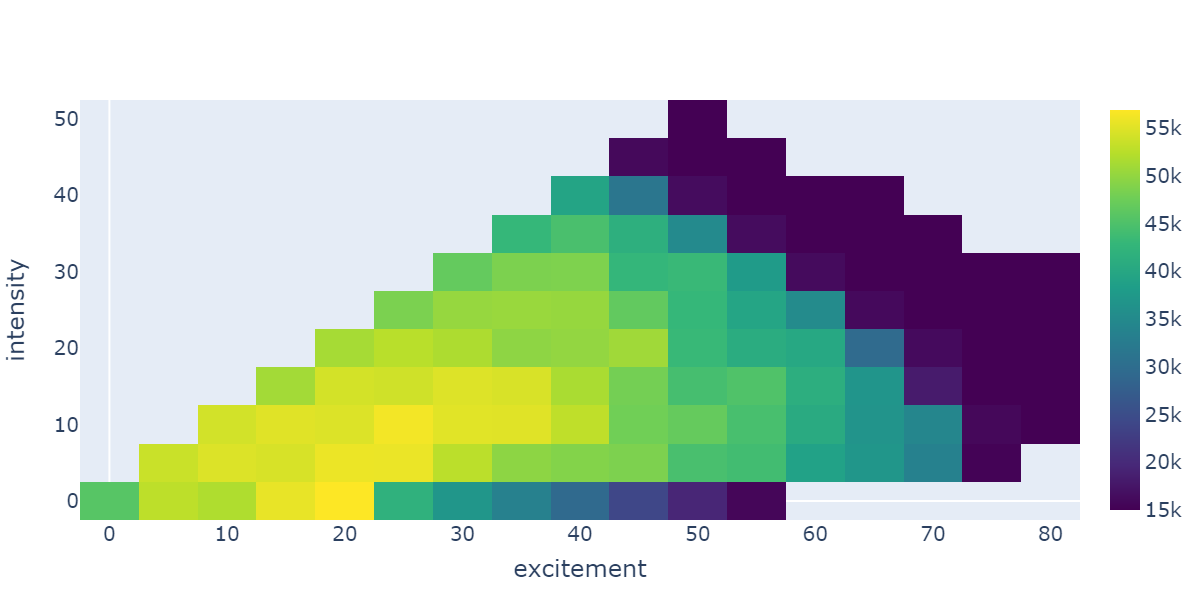}
         \caption{Small initialized, cost disabled park. QD: $\$37,198.09$}
         \label{fig:excitement_intensity_small_no_cost}
     \end{subfigure}
     \centering
     \begin{subfigure}[b]{0.45\linewidth}
         \centering
         \includegraphics[width=\linewidth]{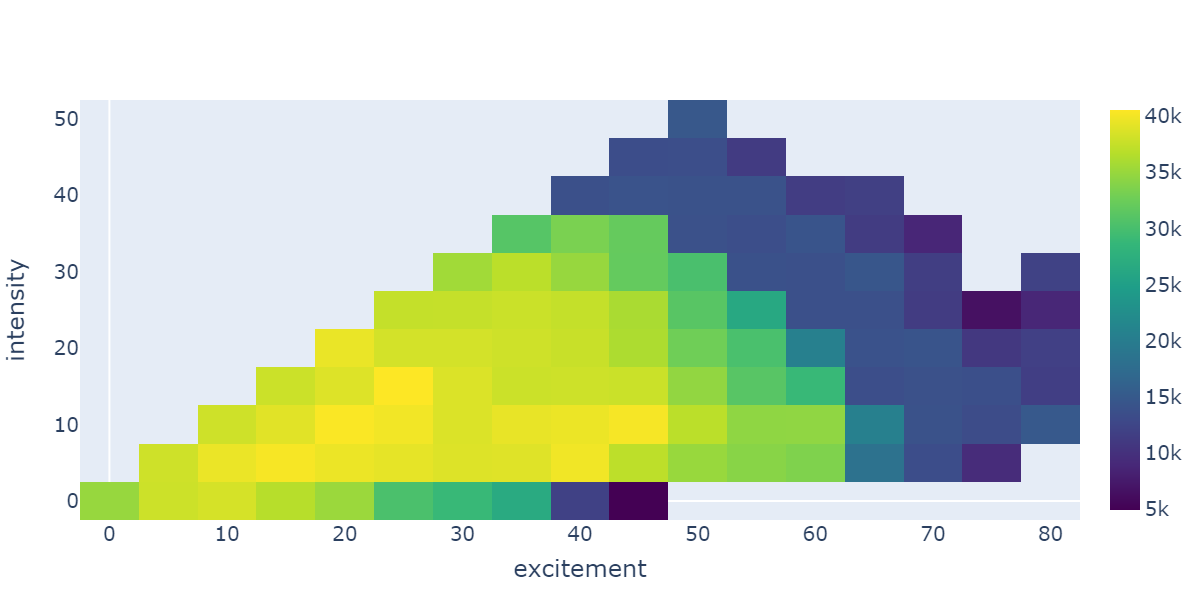}
        \caption{Medium initialized, cost enabled park. QD: $\$26,776.50$}
        \label{fig:excitement_intensity_med_cost}
     \end{subfigure}
     \hfill
     \begin{subfigure}[b]{0.45\linewidth}
         \centering
         \includegraphics[width=\textwidth]{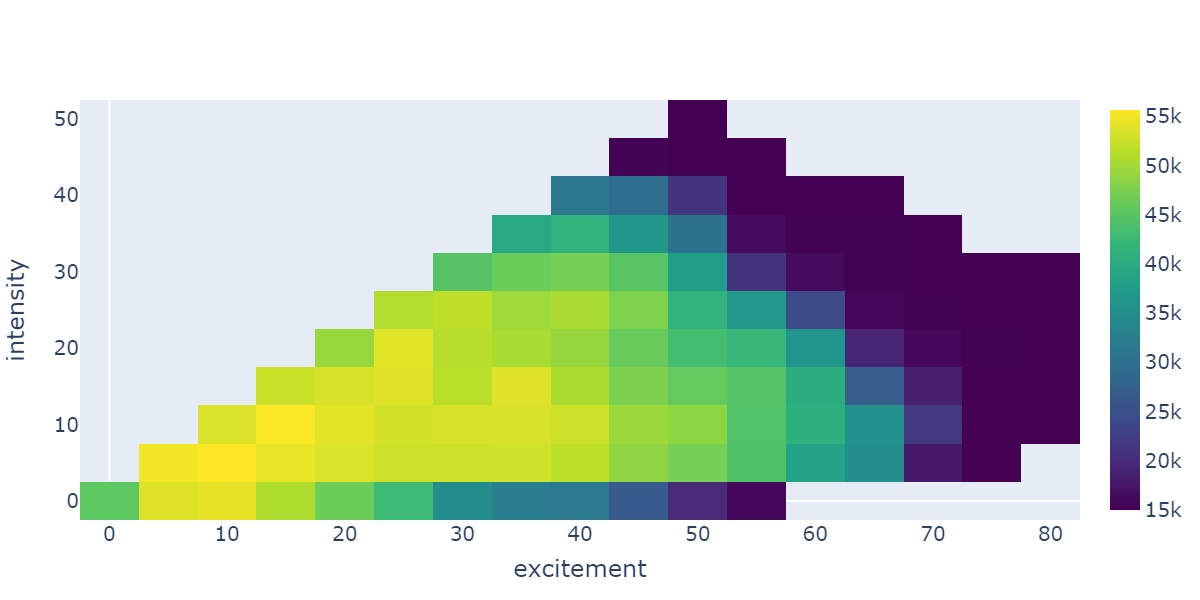}
         \caption{Medium initialized, cost disabled. QD:$\$ 35,975.36$}
         \label{fig:excitement_intensity_med_no_cost}
     \end{subfigure}
    \caption{\textbf{Excitement-Intensity aggregation elite maps.} Brighter colors indicate higher fitness, and each cell is filled with the best elite found over 20 runs. Results indicate that the highest diversity of high-performing park layouts is found when the player starts from scratch and has an unlimited budget.}
    \label{fig:excitement_intensity_agg}
\end{figure*}

\subsection{Happiness Vomit}
\begin{figure*}
     \centering
     \begin{subfigure}[b]{0.45\linewidth}
         \centering
         \includegraphics[width=\linewidth]{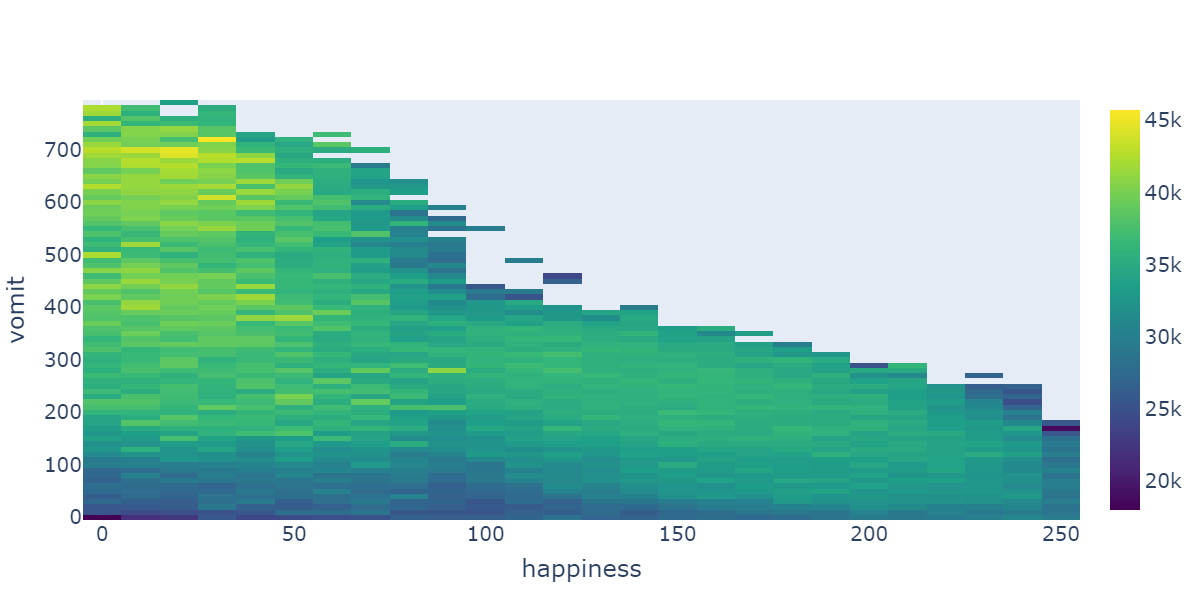}
        \caption{Small initialized, cost enabled park. QD: $\$34,219.07$}
        \label{fig:happiness_vomit_small_cost}
     \end{subfigure}
     \hfill
     \begin{subfigure}[b]{0.45\linewidth}
         \centering
         \includegraphics[width=\textwidth]{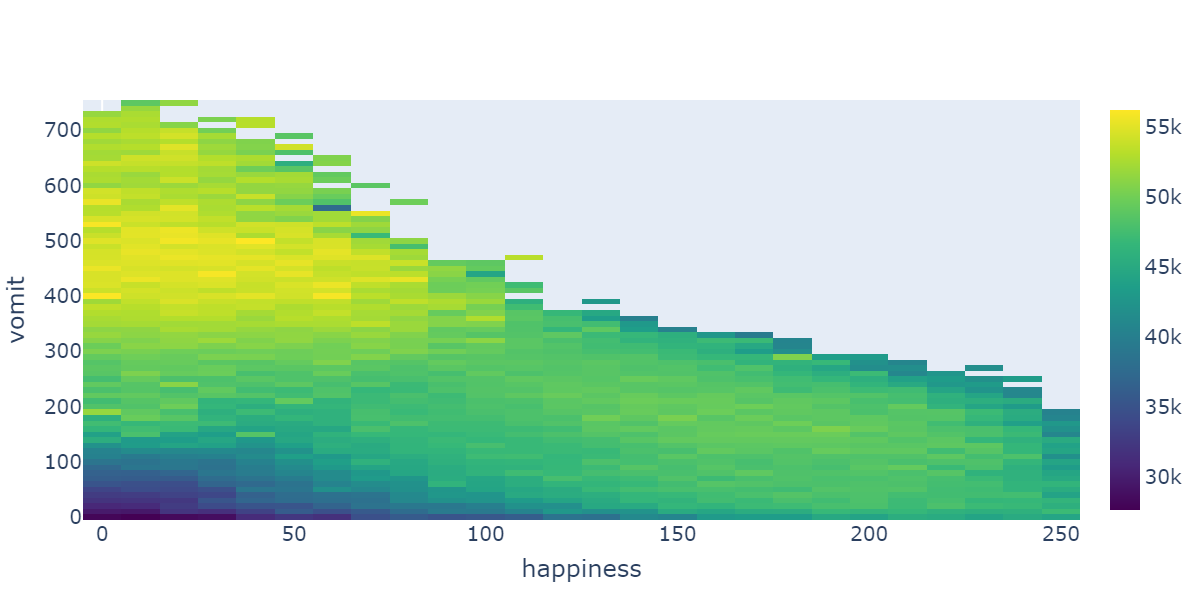}
         \caption{Small initialized, cost disabled park. QD: $\$47,801.68$}
         \label{fig:happiness_vomit_small_nocost}
     \end{subfigure}
     \centering
     \begin{subfigure}[b]{0.45\linewidth}
         \centering
         \includegraphics[width=\linewidth]{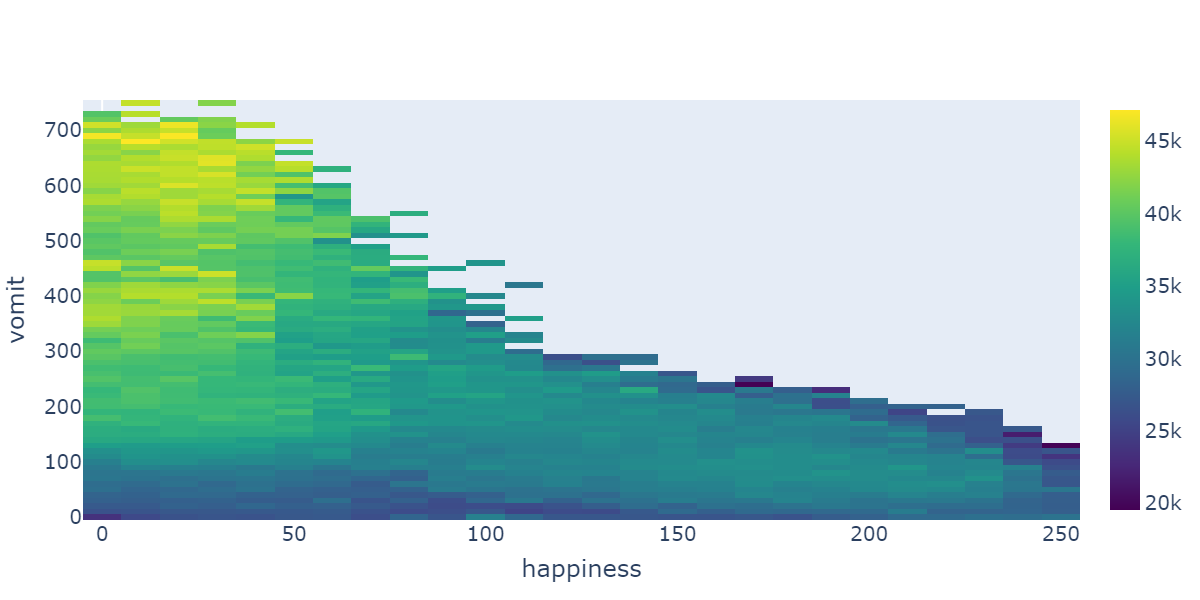}
        \caption{Medium initialized, cost enabled park. QD: $\$34,679.71$}
        \label{fig:happiness_vomit_med_cost}
     \end{subfigure}
     \hfill
     \begin{subfigure}[b]{0.45\linewidth}
         \centering
         \includegraphics[width=\textwidth]{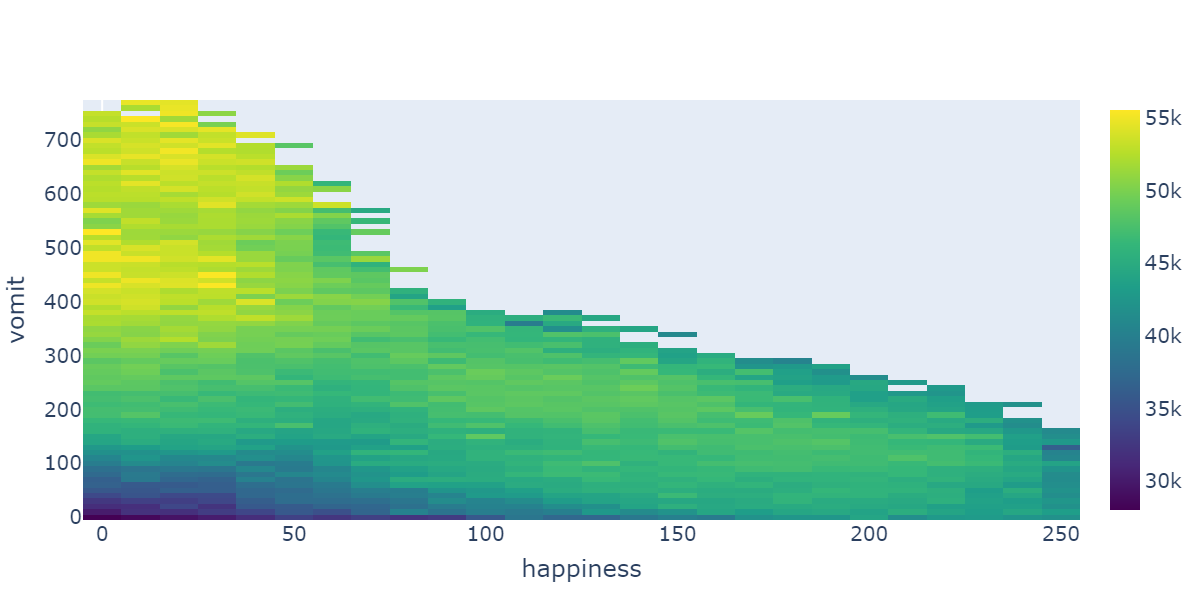}
         \caption{Medium initialized, cost disabled. QD: $\$46,610.39$}
         \label{fig:happiness_vomity_med_nocost}
     \end{subfigure}
    \caption{\textbf{Happiness-Vomit aggregation elite maps.} Brighter colors indicate higher fitness, and each cell is filled with the best elite found over 20 runs. Results again show highest diversity of high-performing rides when the player is given an unlimited budget. Starting not-from-scratch gives a slight performance improvement. The highest fitness rides tend to cluster in the high vomit-low happiness area. Note that the depicted maps have slightly differing color scales.}
    \label{fig:happiness_vomit}
\end{figure*}

\begin{figure}
     \centering
     \begin{subfigure}[b]{0.45\linewidth}
         \centering
         \includegraphics[width=\linewidth]{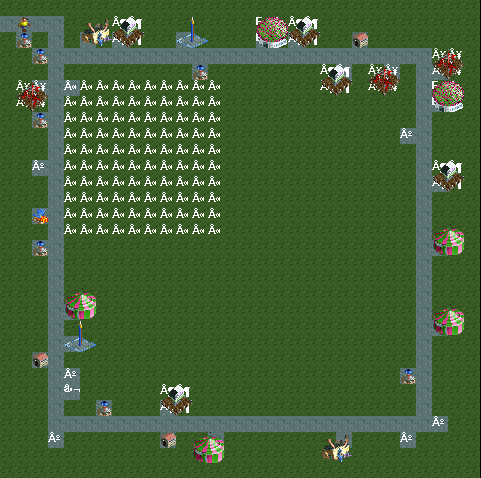}
        \caption{Happiness = 20, Vomit = 540. Fitness = $\$53,455$}
        \label{fig:happiness_vomit_sad}
     \end{subfigure}
     \hfill
     \begin{subfigure}[b]{0.45\linewidth}
         \centering
         \includegraphics[width=\textwidth]{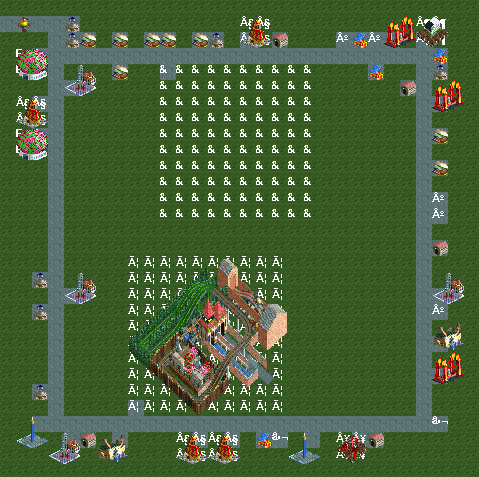}
         \caption{Happiness = $210$, Vomit = $70$. Fitness = $\$46,360$}
         \label{fig:happiness_vomit_happy}
     \end{subfigure}
    \caption{\textbf{Parks from happiness-vomit experiments}, with two different park strategies both leading to high fitness. (a) Has a single coaster, many thrill rides with high nausea ratings, and hardly any concessions. (b) contains a large gentle ride, a high excitement/low nausea coaster, and many concession stands which make peeps happy.}
    \label{fig:happiness_vomit_fitness_compare}
\end{figure}

Happiness and Vomit are both runtime metrics, meaning that the park must be fully simulated to calculate them. Happiness is negatively correlated to vomit: the more vomit that peeps see, the unhappier they will become. Based on the aggregate elite maps (Figure \ref{fig:happiness_vomit}), parks with unhappy guests and high vomit counts have high fitness values. If cost efficient, parks can afford roller-coasters, which allow more revenue and good fitness, though never as high as pure thrill rides. Figure \ref{fig:happiness_vomit_fitness_compare} displays two parks with opposing fitness strategies. The park in Figure \ref{fig:happiness_vomit_sad} contains many thrill rides, one roller-coaster, and almost no other shops or attractions. In contrast, the park in Figure \ref{fig:happiness_vomit_happy}
contains two roller-coasters, a few thrill rides, and a variety of shops and other attractions.

\subsection{Happiness Ride Diversity}
\begin{figure}
     \centering
     \begin{subfigure}[b]{0.45\linewidth}
         \centering
         \includegraphics[width=\linewidth]{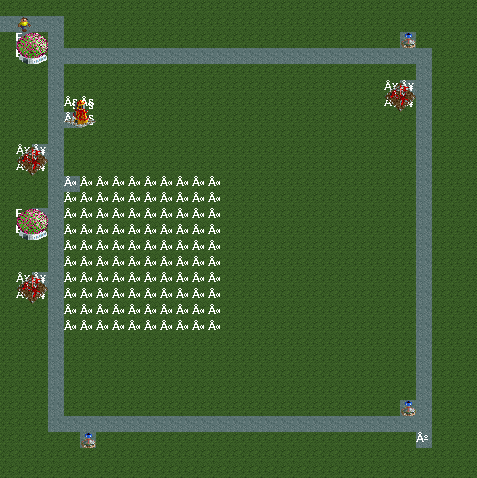}
        \caption{Cost enabled}
        \label{fig:happiness_ridediversity_cost}
     \end{subfigure}
     \hfill
     \begin{subfigure}[b]{0.45\linewidth}
         \centering
         \includegraphics[width=\textwidth]{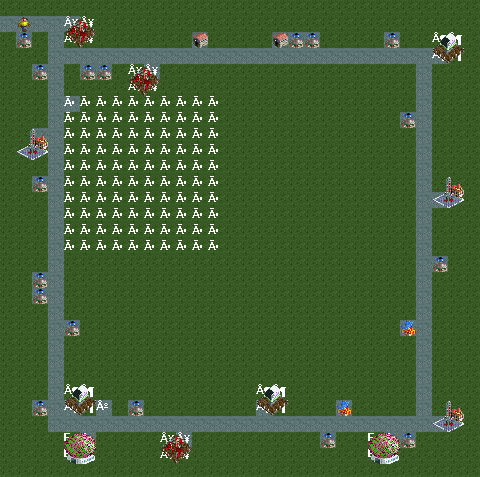}
         \caption{Cost disabled}
         \label{fig:happiness_ridediversity_nocost}
     \end{subfigure}
    \caption{\textbf{Two parks from happiness-ridediversity experiments with different initialization and growth methods}, both with happiness = 20 and ridediversity = 20. Both of these parks were among the highest fitness parks of their experiment type. Thought they have the same dimensional values, they have quite different attractions.}
    \label{fig:happiness_ridediversity_costs}
\end{figure}

\begin{figure}[t]
    \centering
    \includegraphics[width=0.45\linewidth]{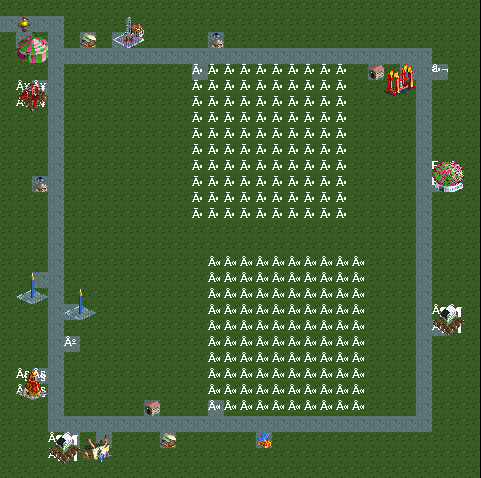}
    \caption{\textbf{A park with happiness and ride diversity values of 250 and 40 respectively}. These dimensional values mean this park is the happiest and most diverse park possible in the framework, at least as discovered by MAPElites. Fitness = $\$25,180$}
    \label{fig:happiness_ridediversity_max}
\end{figure}

Ride Diversity is calculated using the method described in Section \ref{sec:methods-metrics}.
In all four aggregate maps, parks with an average level of ride diversity and low happiness tend to drive more revenue. This is less extreme in the cost disabled evolution runs, most likely due to how cost constraints restrict the number of roller-coasters in the park as mentioned earlier. Figure \ref{fig:happiness_ridediversity_costs} displays two representative maps with cost both enabled and disabled respectively. These parks both have happiness and ride diversity values of $20$ and are among the most fit individuals of their respective runs. Figure \ref{fig:happiness_ridediversity_max} shows a park with the maximum possible values of happiness and diversity. The park has a variety of rides which make peeps happy, and its fitness value is $\$25,180$.

\begin{figure*}
     \centering
     \begin{subfigure}[b]{0.45\linewidth}
         \centering
         \includegraphics[width=\linewidth]{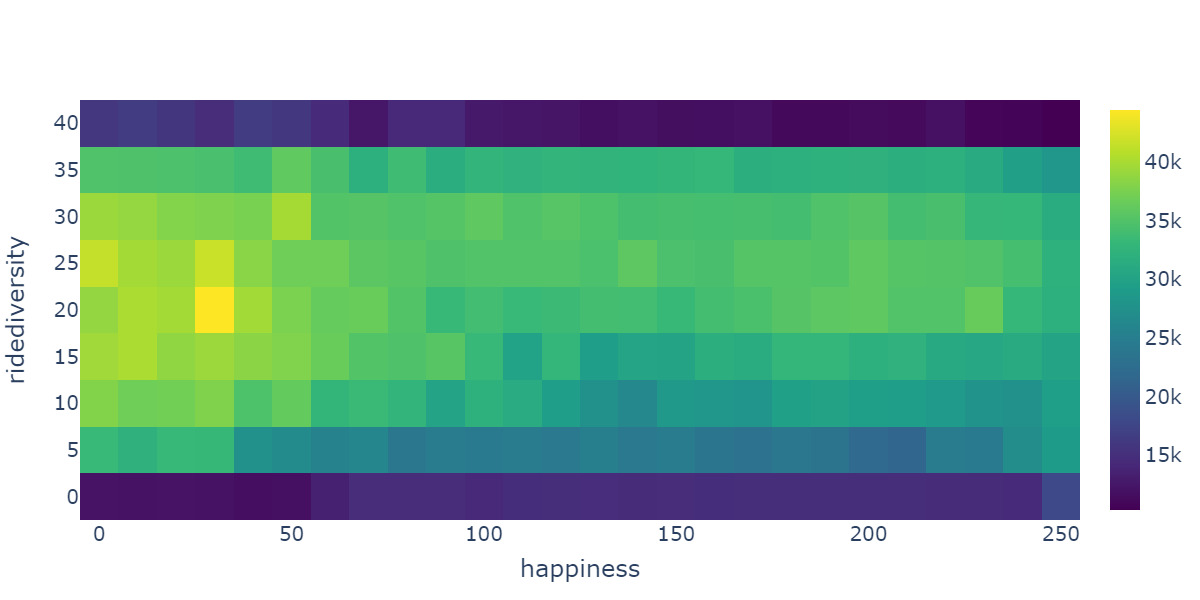}
        \caption{Small initialized, cost enabled. QD: $\$28,634.71$}
        \label{fig:happiness_ridediversity_small_cost}
     \end{subfigure}
     \hfill
     \begin{subfigure}[b]{0.45\linewidth}
         \centering
         \includegraphics[width=\textwidth]{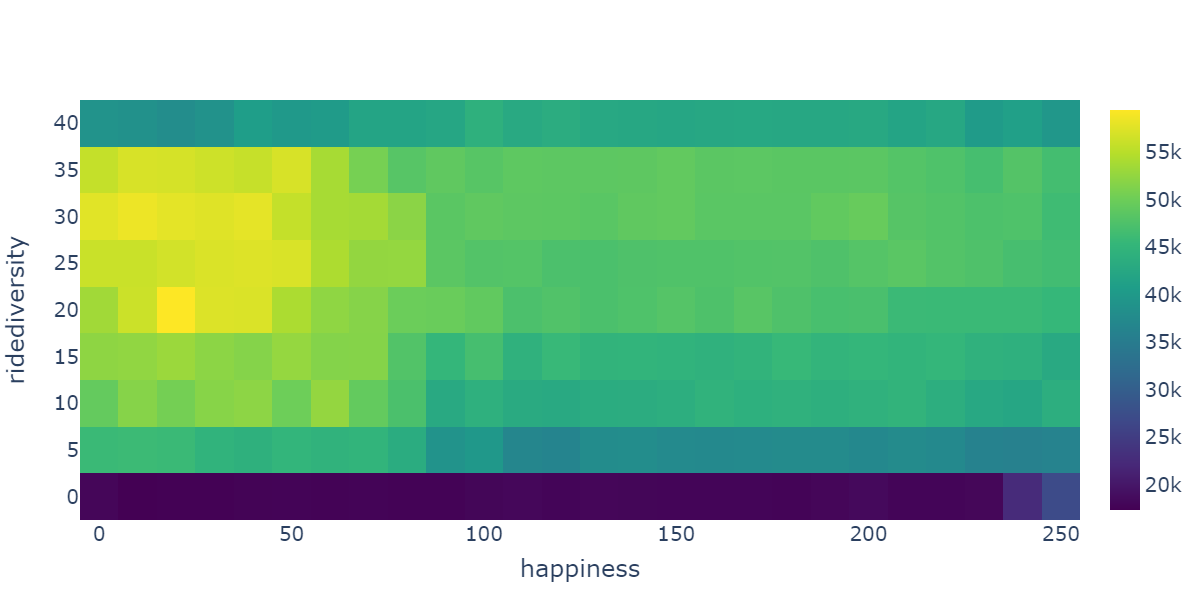}
         \caption{Small initialized, cost disabled. QD: $\$43,829.36$}
         \label{fig:happiness_ridediversity_small_nocost}
     \end{subfigure}
     \centering
     \begin{subfigure}[b]{0.45\linewidth}
         \centering
         \includegraphics[width=\linewidth]{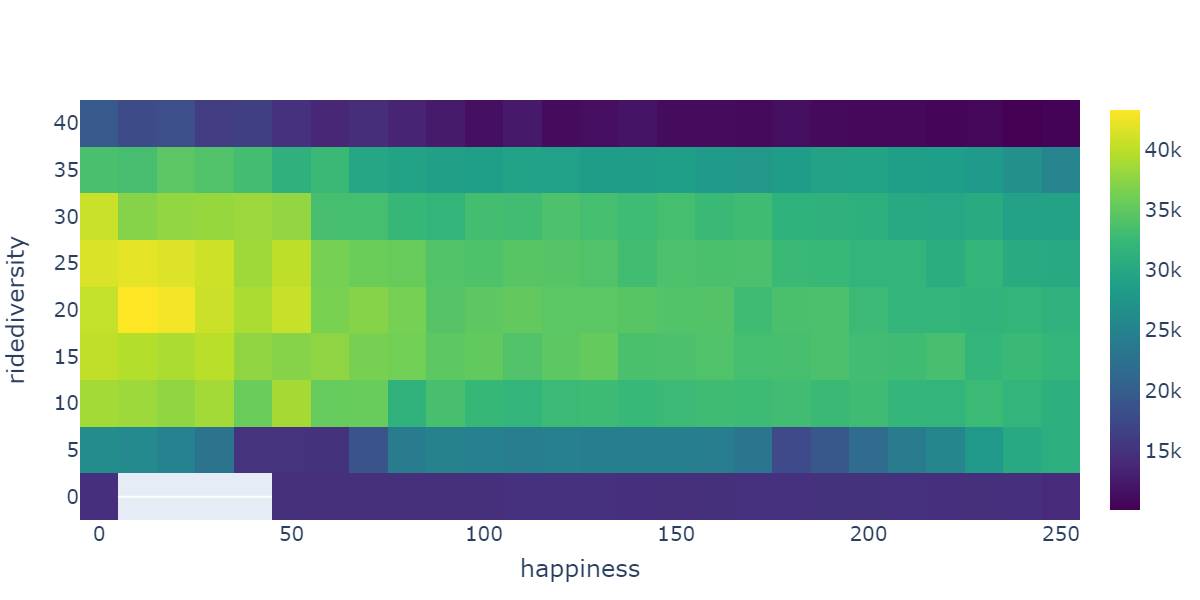}
        \caption{Medium initialized, cost enabled. QD: $\$28,393.72$}
        \label{fig:happiness_ridediversity_med_cost}
     \end{subfigure}
     \hfill
     \begin{subfigure}[b]{0.45\linewidth}
         \centering
         \includegraphics[width=\textwidth]{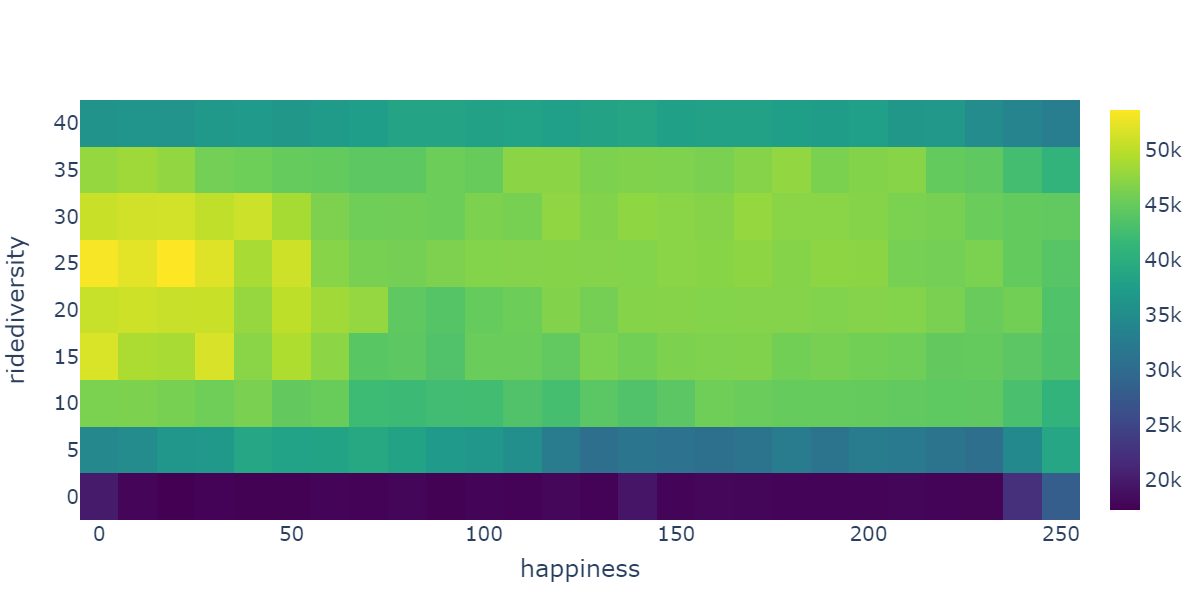}
         \caption{Medium initialized, cost disabled. QD: $\$40,885.04$}
         \label{fig:happiness_ridediversity_med_nocost}
     \end{subfigure}
    \caption{\textbf{Happiness-Ridediversity aggregation elite maps.} Brighter colors indicate higher fitness, and each cell is filled with the best elite found over 20 runs. Results again show highest diversity of high-performing rides when the player starts from scratch and is given an unlimited budget.}
    \label{fig:happiness_ridediversity}
\end{figure*}

\section{Discussion and Conclusion}
RollerCoaster Tycoon is an interesting and unexplored domain for game AI research, a situation this paper aims to rectify by introducing a novel open source domain for evaluating generative systems. The paper focuses on evolutionary algorithms in particular, specifically the MAP-Elites algorithm, but MicroRCT also enables research in general machine learning domains. Preliminary experiments in fact also explored deep reinforcement learning as an alternative to evolution, but with inferior results. 

One question addressed in this paper was whether there is any benefit to starting from a minimal design and adding new components incrementally instead of starting from a system initialized with a moderate amount of random components. Answering this question can give insights into the differences between evolutionary methods that ``complexify'' or augment evolved artifacts over time (e.g. NEAT \cite{stanley:ec02}) and those that do not (e.g. evolution strategies \cite{beyer:nc02}). Exploring this question empirically speaks to the broader theoretical question of exactly how much inspiration evolutionary algorithms must take from nature, wherein a vast diversity of biological life arose from the emergence of the single-celled organism. The results suggest that there is no obvious effect or influence on park evolution with regards to how it is initialized and grown or shrunk over time. Figure \ref{fig:excitement_intensity_bothhigh} demonstrates that even with different initialization and growth methods, evolution can reach the same conclusion. This does not mean that these growth methods will generate completely identical parks. There may be other dimensions not studied in this paper in which differences are more pronounced.

The second question addressed in this paper was about the effects of resource constraints on MAP-Elites. This algorithm has emerged as a recently popular tool for generative design in a variety of application domains including robotics and games, with the majority of applications existing entirely in simulation. Besides the insights gained in terms of evolutionary theory, for practical purposes it is imperative that we understand how MAP-Elites performs in resource-limited domains if it is to be applied in more real-world domains with physical constraints. Cost constraints have an obvious effect of the fitness ceiling of the evolved parks, as well as the overall quality diversity score of the elite maps. In general, parks with cost constraints are unable to build more than one roller-coaster, if any. Although more expensive than thrill rides, roller-coasters generate more revenue. Therefore, the more there are in the park, the more money that park can make.

Additionally, answering these two questions has implications for game design. If a designer wants to make a game more open-ended, they might choose to force the player to start from scratch. Adding resource constraints limits the number of possible designs, but also increases the difficulty of the game by restricting the design space and forcing the player to possibly think more creatively when there isn't an obviously optimal way to play.


\bibliographystyle{IEEEtran}
\bibliography{IEEEfull,bibliography}

\begin{thebibliography}{10}
\providecommand{\url}[1]{#1}
\csname url@samestyle\endcsname
\providecommand{\newblock}{\relax}
\providecommand{\bibinfo}[2]{#2}
\providecommand{\BIBentrySTDinterwordspacing}{\spaceskip=0pt\relax}
\providecommand{\BIBentryALTinterwordstretchfactor}{4}
\providecommand{\BIBentryALTinterwordspacing}{\spaceskip=\fontdimen2\font plus
\BIBentryALTinterwordstretchfactor\fontdimen3\font minus
  \fontdimen4\font\relax}
\providecommand{\BIBforeignlanguage}[2]{{%
\expandafter\ifx\csname l@#1\endcsname\relax
\typeout{** WARNING: IEEEtran.bst: No hyphenation pattern has been}%
\typeout{** loaded for the language `#1'. Using the pattern for}%
\typeout{** the default language instead.}%
\else
\language=\csname l@#1\endcsname
\fi
#2}}
\providecommand{\BIBdecl}{\relax}
\BIBdecl

\bibitem{sawyerFAQ}
``...about {C}hris {S}awyer \& game development,''
  \url{http://www.chrissawyergames.com/faq3.htm}, accessed: March 24, 2021.

\bibitem{cully:nature15}
A.~Cully, J.~Clune, D.~Tarapore, and J.-B. Mouret, ``Robots that can adapt like
  animals,'' \emph{Nature}, vol. 521, no. 7553, pp. 503--507, 2015.

\bibitem{pugh:frontiers16}
J.~K. Pugh, L.~B. Soros, and K.~O. Stanley, ``Quality diversity: A new frontier
  for evolutionary computation,'' \emph{Frontiers in Robotics and AI}, vol.~3,
  p.~40, 2016.

\bibitem{QDio}
``Quality-diversity optimisation algorithms,''
  \url{https://quality-diversity.github.io/papers}, accessed: March 24, 2021.

\bibitem{khalifa:gecco18}
A.~Khalifa, S.~Lee, A.~Nealen, and J.~Togelius, ``Talakat: Bullet hell
  generation through constrained map-elites,'' in \emph{Proceedings of The
  Genetic and Evolutionary Computation Conference}, 2018, pp. 1047--1054.

\bibitem{gravina2019procedural}
D.~Gravina, A.~Khalifa, A.~Liapis, J.~Togelius, and G.~N. Yannakakis,
  ``Procedural content generation through quality diversity,'' \emph{arXiv
  preprint arXiv:1907.04053}, 2019.

\bibitem{alvarez2019empowering}
A.~Alvarez, S.~Dahlskog, J.~Font, and J.~Togelius, ``Empowering quality
  diversity in dungeon design with interactive constrained map-elites,''
  \emph{arXiv preprint arXiv:1906.05175}, 2019.

\bibitem{charity2020mech}
M.~Charity, M.~C. Green, A.~Khalifa, and J.~Togelius, ``Mech-elites:
  Illuminating the mechanic space of gvg-ai,'' in \emph{International
  Conference on the Foundations of Digital Games}, 2020, pp. 1--10.

\bibitem{cully:ToEC17}
A.~Cully and Y.~Demiris, ``Quality and diversity optimization: A unifying
  modular framework,'' \emph{IEEE Transactions on Evolutionary Computation},
  vol.~22, no.~2, pp. 245--259, 2017.

\bibitem{lehman:gecco2011}
J.~Lehman and K.~O. Stanley, ``Evolving a diversity of virtual creatures
  through novelty search and local competition,'' in \emph{Proceedings of the
  13th annual conference on Genetic and evolutionary computation}, 2011, pp.
  211--218.

\bibitem{cully:gecco19}
A.~Cully, ``Autonomous skill discovery with quality-diversity and unsupervised
  descriptors,'' in \emph{Proceedings of the Genetic and Evolutionary
  Computation Conference}, 2019, pp. 81--89.

\bibitem{bossens:gecco20}
D.~M. Bossens, J.-B. Mouret, and D.~Tarapore, ``Learning behaviour-performance
  maps with meta-evolution,'' in \emph{Proceedings of the 2020 Genetic and
  Evolutionary Computation Conference}, 2020, pp. 49--57.

\bibitem{rios:sb09}
L.~H.~O. {Rios} and L.~{Chaimowicz}, ``trains: An artificial inteligence for
  openttd,'' in \emph{2009 VIII Brazilian Symposium on Games and Digital
  Entertainment}, 2009, pp. 52--63.

\bibitem{wisniewski:thesis}
M.~Wisniewski and W.~Carsten, ``Artificial intelligence for the openttd game,''
  Ph.D. dissertation, Kongens Lyngby: Technical University of Denmark, 2011.

\bibitem{bijlsma:thesis}
F.~Bijlsma, ``Evolving dynamic ai opponents for openttd using dynamic scripting
  and grammatical evolution,'' Master's thesis, 2014.

\bibitem{lakomy:thesis}
O.~Lakom{\`y}, ``Railroad network planning in open transport tycoon deluxe,''
  2020.

\bibitem{katreniakova:thesis}
T.~Katreniakov{\'a}, ``Konfigurovateln{\'a} um{\v{e}}l{\'a} inteligence pro
  openttd,'' Ph.D. dissertation, Masarykova univerzita, Fakulta informatiky,
  2020.

\bibitem{konijnendijk:thesis}
G.~Konijnendijk, ``Mcts in openttd,'' Ph.D. dissertation, Maastricht Universit,
  2015.

\bibitem{earle2020using}
S.~Earle, ``Using fractal neural networks to play simcity 1 and conway's game
  of life at variable scales,'' \emph{arXiv preprint arXiv:2002.03896}, 2020.

\bibitem{charity:aiide20}
M.~Charity, D.~Rajesh, R.~Ombok, and L.~Soros, ``Say “sul sul!” to simsim,
  a sims-inspired platform for sandbox game ai,'' in \emph{Proceedings of the
  AAAI Conference on Artificial Intelligence and Interactive Digital
  Entertainment}, vol.~16, no.~1, 2020, pp. 182--188.

\bibitem{soros:alife14}
L.~Soros and K.~Stanley, ``Identifying necessary conditions for open-ended
  evolution through the artificial life world of chromaria,'' in
  \emph{Artificial Life Conference Proceedings 14}.\hskip 1em plus 0.5em minus
  0.4em\relax MIT Press, 2014, pp. 793--800.

\bibitem{brant:gecco17}
J.~C. Brant and K.~O. Stanley, ``Minimal criterion coevolution: a new approach
  to open-ended search,'' in \emph{Proceedings of the Genetic and Evolutionary
  Computation Conference}, 2017, pp. 67--74.

\bibitem{russell2002artificial}
S.~Russell and P.~Norvig, ``Artificial intelligence: a modern approach,'' 2002.

\bibitem{pugh:thesis}
J.~Pugh, ``Quality diversity: Harnessing evolution to generate a diversity of
  high-performing solutions,'' 2019.

\bibitem{stanley:ec02}
K.~O. Stanley and R.~Miikkulainen, ``Evolving neural networks through
  augmenting topologies,'' \emph{Evolutionary computation}, vol.~10, no.~2, pp.
  99--127, 2002.

\bibitem{beyer:nc02}
H.-G. Beyer and H.-P. Schwefel, ``Evolution strategies--a comprehensive
  introduction,'' \emph{Natural computing}, vol.~1, no.~1, pp. 3--52, 2002.

\end{thebibliography}

\end{document}